# ADJUSTING FOR BIAS WITH PROCEDURAL DATA

**Shesh Narayan Gupta, Nicholas Bear Brown**
Northeastern University, Boston, MA
02120, USA
{gupta.shesh@northeastern.edu, ni.brown@northeastern.edu}

## ABSTRACT

3D software's are now capable of producing highly realistic images that look nearly indistinguishable from the real images. This raises the question: can real datasets be enhanced with 3D rendered data? We investigate this question. In this paper we demonstrate the use of 3D rendered data, procedural, data for the adjustment of bias in image datasets. We perform error analysis of images of animals which shows that the misclassification of some animal breeds is largely a data issue. We then create procedural images of the poorly classified breeds and that model further trained on procedural data can better classify poorly performing breeds on real data. We believe that this approach can be used for the enhancement of visual data for any underrepresented group, including rare diseases, or any data bias potentially improving the accuracy and fairness of models. We find that the resulting representations rival or even out- perform those learned directly from real data, but that good performance requires care in the 3D rendered/procedural data generation. 3D image dataset can be viewed as a compressed and organized copy of a real dataset, and we envision a future *[19]* where more and more procedural data proliferate while datasets become increasingly unwieldy, missing, or private. This paper suggests several techniques for dealing with visual representation learning in such a future. Code is available on our project page https://github.com/aiskunks/AI_Research/tree/main/Adjustin_bias_using_procedural_data.

# 1 INTRODUCTION

The last few years have seen great progress in the diversity and quality of 3D rendering and modeling software like Unreal Engine, Blender, SideFX Houdini and many more. 3D rendering software can produce realistic samples of images and video that is shot at any angle, in any lighting condition, in any pose, and in any environment. This raises an intriguing possibility: can we now enhance real data with 3D rendered data?

## Procedural Data can help with computer vision data issues

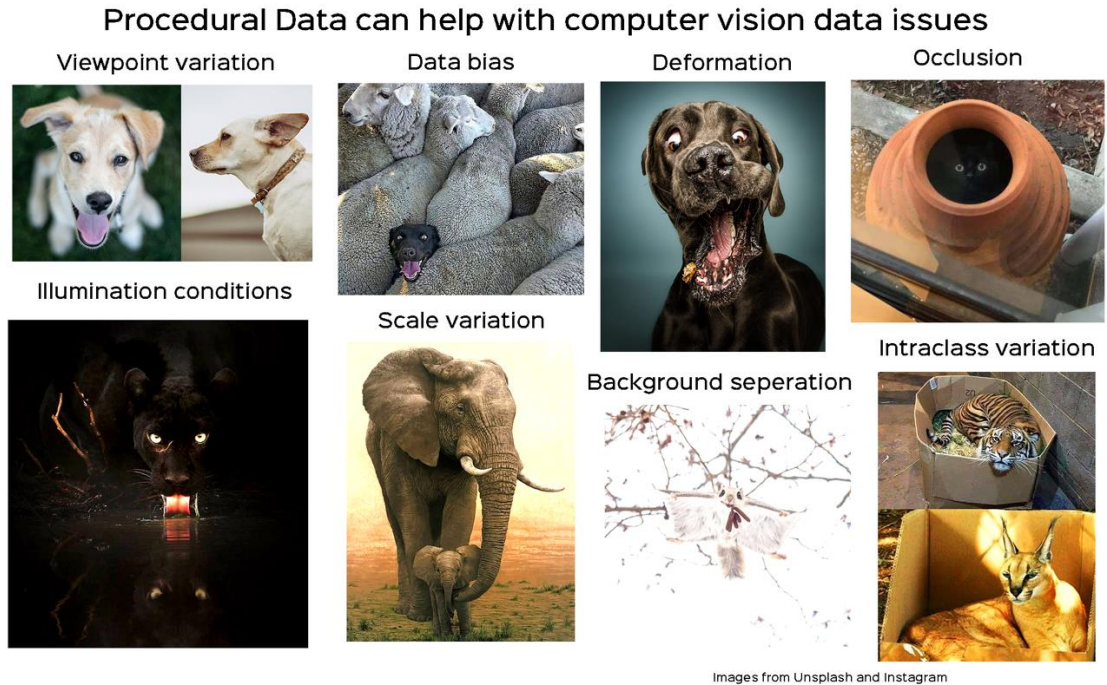

*Figure 1: Procedural data, 3D rendered data from game and 3D engines, has the potential to address some of computer visions most fundamental data issues*

If so, there would be immediate advantages. These 3D images are highly accurate have better lighting, angles and are compressed compared to the real datasets they represent and therefore easier to share and store. 3D image data also circumvents some of the concerns around privacy and usage rights that limit the distribution of real datasets [6], and rendered images and videos can further be edited to censor sensitive attributes [7]. Perhaps because of these advantages, it is becoming increasingly common for 3D images and 3D image data to be shared online freely for anyone to use. This approach has been taken by individuals who may not have the resources or intellectual property rights to release the original data and to counter for rights/political issues.

Our work therefore targets a problem setting that has received little prior attention: given access to a 3D and game engines, and access to the real dataset with limited number of images and of low quality, can we learn effective visual representations and improve the accuracy of CNN classifiers to correctly classify the image class to which it belongs and in turn remove any unwanted bias in the model?

To this end, we provide an exploratory study of adjusting the bias in real data in the setting of 3D image data taken from various 3D rendering software's: we analyzed what is the accuracy of CNN classifier when trained only with real animal images in 5 categories namely bear, sheep, dog, elephant and horse, how well they work, and how they can be improved to make use of the high-quality 3D image data generated by 3D software. Figure.2 lays out the framework we study:

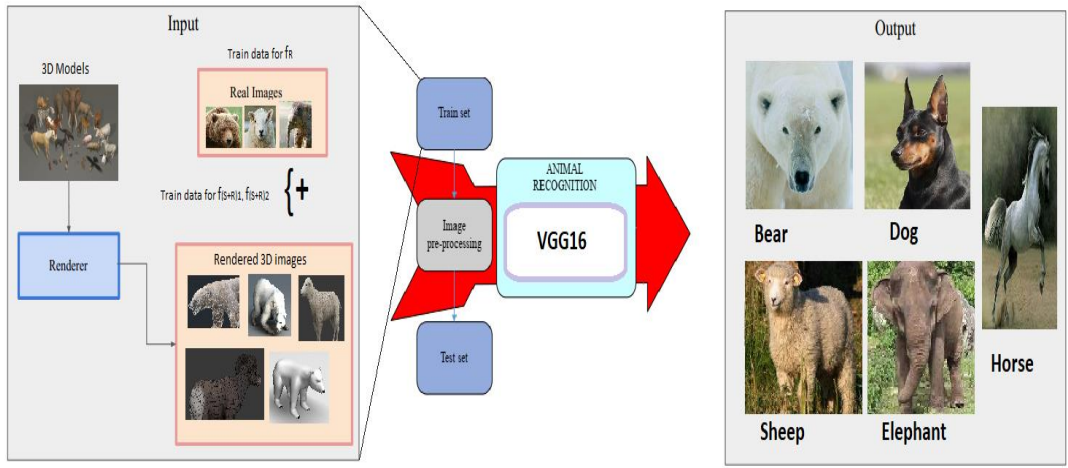

**Figure 2:** *Study framework*

We took around 10,000 real images of 5 animal categories namely bear, dog, sheep, horse and elephant and trained a CNN to classify the unseen animal images to its correct categories. We then studied the accuracy of the CNN which is the misclassification rates of each animal category and ask under what conditions they can lead to enhanced accuracy. We then compared these results with the accuracy results of the same CNN but trained on the data after adding around 200 3D rendered images of bear category. Further, we studied the accuracy result of the same CNN trained on data which contained 3D rendered images of bear and sheep category (3D rendered images of sheep were added on top of 3D rendered images of bear) to identify if we can adjust the misclassification that is if we can adjust the bias using 3D rendered images and increase the overall accuracy of the CNN model. We asked: whether adding 3D rendered images helped to adjust the bias in data and if the model now can better classify poorly performing breeds on real data? Our main findings are:

1. 3D image data i.e., the 3D rendered images can be used to adjust the bias in the data but properly matching the distribution of the real data with the 3D data can affect the results and needs further study.

2. These 3D image data can be combined with the real-world data to achieve better accuracy and lower misclassification in the CNN model than on the real-world data alone.

3. The lighting, quality and angles of the 3D rendered images plays a key role in adjusting the bias. Tools that analyze the real data and suggest which poses and lighting conditions are needed for the 3D data is a crucial next step in this work.

4. 3D rendering software's can potentially produce an unbounded number of samples to train on; we find that performance improves by training on more samples.

5. Adding 3D image data may decrease the accuracy, we believe this happens when the rendered images are less representees of the real image distributions, but this requires further study. Tools that analyze the real data and suggest the parameters of the 3D images is a clear next step in this work.

Below tables summarize the accuracies of the model obtained by testing on 500 bootstrap samples containing 200 images of each animal.

**Table 1:** *Confidence interval and accuracies of model $f_R$, bootstrapped over 500 times*

| Model | $f_R$ | | | |
|---|---|---|---|---|
| Animal/Stats | min | mean | max | Confidence interval (with 95% confidence) |
| bear | 0.435 | 0.535 | 0.635 | (0.5316845879502172,  0.5379154120497824) |
| dog | 0.78 | 0.87 | 0.94 | (0.86852343155255,  0.8727165684474524) |
| elephant | 0.705 | 0.802 | 0.89 | (0.7991678845343991,  0.8041521154655981) |
| horse | 0.585 | 0.706 | 0.8 | (0.7029347731939645,  0.7083852268060363) |
| sheep | 0.575 | 0.683 | 0.79 | (0.6797492149265962,  0.6856707850734041) |

**Table 2:** *Confidence interval and accuracies of model $f_{(S+R)1}$, bootstrapped over 500 times*

| Model | $f_{(S+R)1}$ | | | |
|---|---|---|---|---|
| Animal/Stats | min | mean | max | Confidence interval (with 95% confidence) |
| bear | 0.56 | 0.648 | 0.745 | (0.6454438770018157,  0.6513561229981849) |
| dog | 0.72 | 0.823 | 0.9 | (0.8205027976389079,  0.8253572023610929) |
| elephant | 0.725 | 0.813 | 0.89 | (0.8100958715575248,  0.8150841284424766) |
| horse | 0.58 | 0.664 | 0.75 | (0.6612672056725515,  0.6668927943274474) |
| sheep | 0.645 | 0.742 | 0.82 | (0.7394166283453301,  0.7446033716546727) |

**Table 3:** *Confidence interval and accuracies of model $f_{(S+R)2}$, bootstrapped over 500 times*

| Model | $f_{(S+R)2}$ | | | |
|---|---|---|---|---|
| Animal/Stats | min | mean | max | Confidence interval (with 95% confidence) |
| bear | 0.56 | 0.649 | 0.735 | (0.6461913719795908,  0.6518886280204089) |
| dog | 0.79 | 0.864 | 0.945 | (0.8615758814693575,  0.8659441185306442) |
| elephant | 0.715 | 0.808 | 0.905 | (0.8054984209403556,  0.8101615790596438) |
| horse | 0.565 | 0.653 | 0.75 | (0.6495009443415476,  0.6555390556584522) |
| sheep | 0.64 | 0.756 | 0.855 | (0.7537195756306722,  0.7589404243693271) |

## 2 RELATED WORK

**Learning from 3D image data.** Using 3D image data has been a prominent method for learning in different domains of science and engineering, with different goals including privacy-preservation and alternative sample generation *[6,8,9,10]*. In computer vision, 3D image data has been extensively used as a source for training models, for example in semantic segmentation *[12]*, human pose estimation *[11]* or self-supervised multitask learning *[13]*. In most prior work, the 3D image data comes from a traditional simulation pipeline, e.g., via rendering a 3D world with a graphics engine, or recently a noise generator *[14]*.

**Animal Parsing.** Though there exist large scale datasets containing animals for classification, detection, and instance segmentation, these datasets only cover a tiny portion of animal species in the world. Due to the lack of annotations, 3D image data has been widely used to address the problem *[1, 2, 3, 4]*. Similar to SMPL models *[5]* for humans, *[4]* proposes a method to learn articulated SMAL shape models for animals. Later, *[3]* extracts more 3D shape details and is able to model new species. Unfortunately, these methods are built on manually extracted silhouettes and key point annotations. Recently, *[2]* proposes to copy texture from real animals and predicts 3D mesh of animals in an end-to-end manner. Most related to our method is *[1]*, where authors propose a method to estimate animal poses on real images using synthetic silhouettes. Different from *[3]* which focuses on estimating animal poses, our strategy is to use 3D image data to adjust the bias in model classification for underrepresented animal breeds.

## 3 METHOD

We consider the problem under unsupervised framework with two datasets. We name our real animal image dataset ($\mathbf{X_R}$) and 3D rendered animal image dataset ($\mathbf{Xs}$) as the source dataset and real world animal images (different from the real world animal source dataset) as the target dataset $\mathbf{Xt}$. The goal is to trian a. model $\mathbf{f}$ to predict categories/label for the target data $\mathbf{X_t}$. We simply start with training our first model $\mathbf{f_R}$ with real world data ($\mathbf{X_R}$) using Vgg16 architecture and transfer learning. Then we used the trained model to perform the prediction on the test dataset $\mathbf{X_t}$. We then added our 3D image dataset and trained the model $\mathbf{f_{(S+R)1}}, \mathbf{f_{(S+R)2}}$ in the same setting but on dataset that contained both 3D rendered images and real images. The 3D image data however was added one animal category each time for two animal categories, so at the end, we had 3 models: $\mathbf{f_R}$ trained model with only real images of 5 animal categories, $\mathbf{f_{(S+R)1}}$ trained model with real images and 3D rendered images of bear and $\mathbf{f_{(S+R)2}}$ trained model with real images and 3D rendered images of bear and sheep. Finally, we compared accuracy of all the models and studied the accuracy on all animal categories. Table 4 displays the comparison results of all the three models

**Table 4:** *Comparison result for the mean of 3 model accuracies, bootstrapped over 500 times*

| Animal/Model | $f_R$ | $f_{(S+R)1}$ | $f_{(S+R)2}$ |
|---|---|---|---|
| bear | 0.5348 | 0.6484 | 0.64904 |
| dog | 0.87062 | 0.82293 | 0.86376 |
| elephant | 0.80166 | 0.81259 | 0.80783 |
| horse | 0.70566 | 0.66408 | 0.65252 |
| sheep | 0.68271 | 0.74201 | 0.75633 |

Figure 3 represents the visual comparison of the results

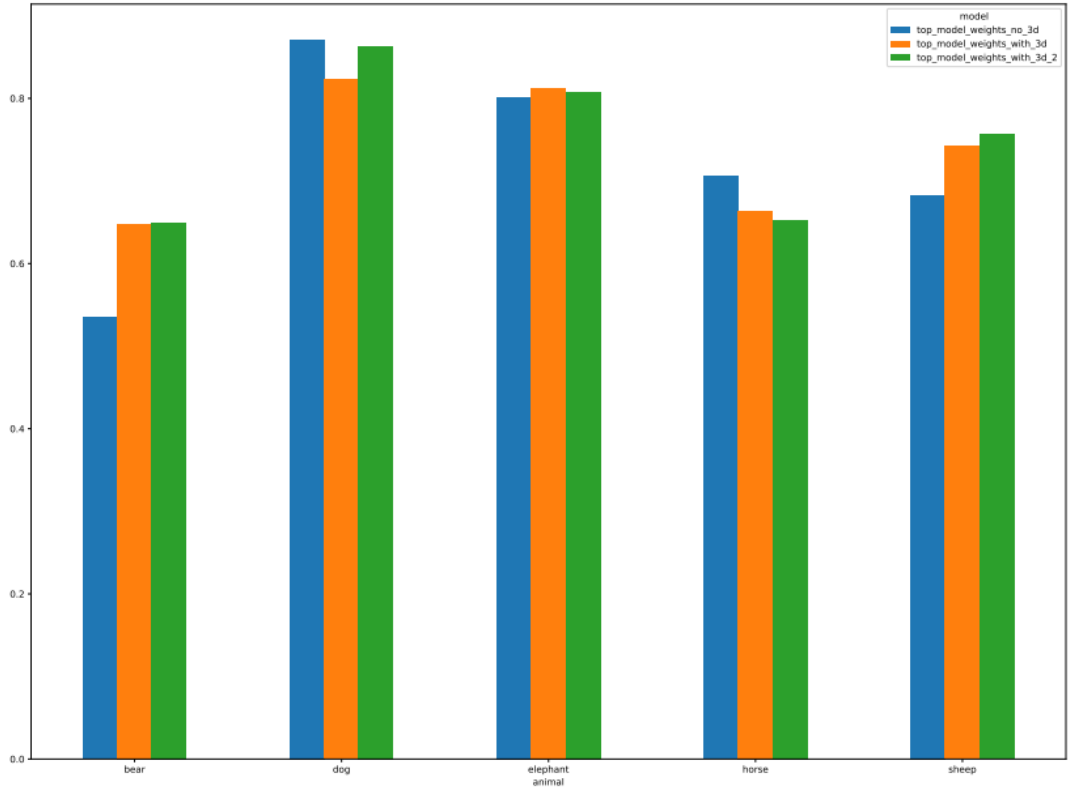

***Figure 3:*** *Comparison result for the 3 models*

**\*Note:** $f_R$ = top_model_weights_no_3d (blue), $f_{(S+R)1}$ = top_model_weights_with_3d (orange), and

$f_{(S+R)2}$ = top_model_weights_with_3d_2 (green)

## 4  EXPERIMENTS

We experiment on a single CNN with 3 training datasets as described in section 3. We evaluate our model on the test image set which contained 1000 images belonging to 5 animal categories under study.

**Network Architecture**. We use Convolutional Neural Network with Leaky ReLU activation function and VGG16 architecture for our model. The initial step aims at creation of features with VGG16 model. Application of Image Processing along with Loading, Testing, Training, and Validating the dataset before the training step helps to remove the noise, obstacles, distortion and dirt from the images. The next step uses Convolutional Neural Network along with Leaky ReLU to train the model to accurately and precisely classify animal classes. In order to avoid the problem of Dying ReLU, where some ReLU neurons essentially die for all inputs and remain inactive no matter what input is supplied, here no gradient flows and if large number of dead neurons are there in a neural. network its performance is affected. To resolve this issue, we make use of what is called Leaky ReLU, where slope is changed left of x=0 and thus causing a leak and extending the range of ReLU. We trained the model for 100 ephocs and used RMSPROP optimizer wih the learning rate of $1e^{-4}$. After training the model, we graph the model's training and validation accuracy and loss to have insights about how well the model is trained. The accuracy and loss graphs for all the 3 models under study are provided in the [Appendix](#).

**3D image datasets**. We explain the details of our data generation parameters as follows. The virtual camera has a resolution of 640×480 and field of view of 90. We generated 200 images with random texture for bear and sheep animal categories.

**Real image datasets**. We explain the details of our data generation parameters as follows. The real images of the animals in the categories were taken from Imges.cv ( https://images.cv/ ) website and Kaggle animal dataset ( https://www.kaggle.com/datasets/alessiocorrado99/animals10 ). We took around 10000 images in 5 animal categories. We split the training set and validation set with a ratio of 4:1, resulting in 8,000 images for training and 2,000 for validation.

## 5 CONCLUSION

We investigated how to reduce the bias in the dataset. 3D rendering software's make it possible to generate multiple views of similar image content with variety of angles, lighting and poses. These 3D rendered images can be used for training the model along with the real-world images to remove the bias and improve the accuracy. Our results demonstrate that leveraging 3D image data can improve accuracy beyond learning from real world images alone. However, this is not as simple as adding more pictures of bears, if a model is doing poorly on bear classification. Adding large quantities of 3D image data sometimes undermines the accuracy of categories which have some semblance amongst each other.

Tools need to be developed which suggest what types of 3D images need to be rendered is crucial. The next step in this research is to study why adding 3D image helps and when it can hurt so that 3D images can be generated that are specifically for that data.

## 6 ETHICS STATEMENT

This work studies how to adjust the bias and learn useful image representations given data generated from 3D rendering software's as opposed to real data. This framework can further provide several societal advantages currently faced in real datasets, including protecting the privacy and usage rights of real images [6], or removing sensitive attributes [7] and therefore learning fairer representations. At the same time, learning representations from 3D rendered images brings several challenges and risks, that need to be addressed. Rendered images can in some cases have the angles, lighting which are not the accurate representation of the real-world image. They can also amplify biases in datasets [15] which can lead to negative societal impacts if they are not audited [16,17,18], or they are used in the wrong contexts. Making good use of the 3D image data will require addressing the above issues by studying mitigation strategies, as well as methods to audit implicit rendered images.

## ACKNOWLEDGMENTS

Author Nicholas Bear Brown for their support. Special thanks to Shaunit Narayan Gupta for motivating me and encouraging me to continue work at nights as well.

Proceedings of the AAAI/ACM Conference on AI, Ethics, and Society, pp. 145–151, 2020a.

APPENDIX

**$f_{(S+R)2}$ loss and accuracy**

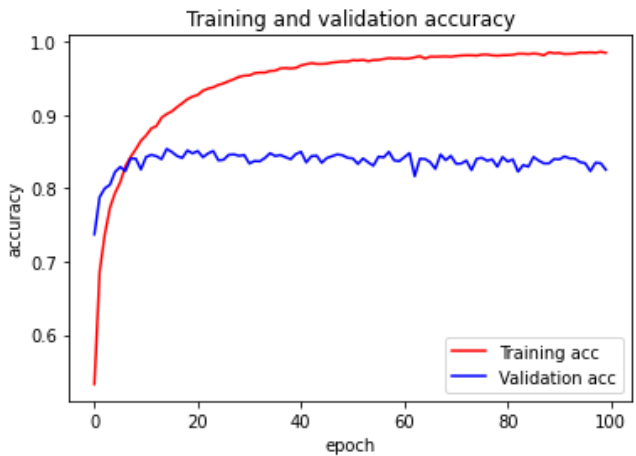

**Figure 4:** *Model Accuracy*

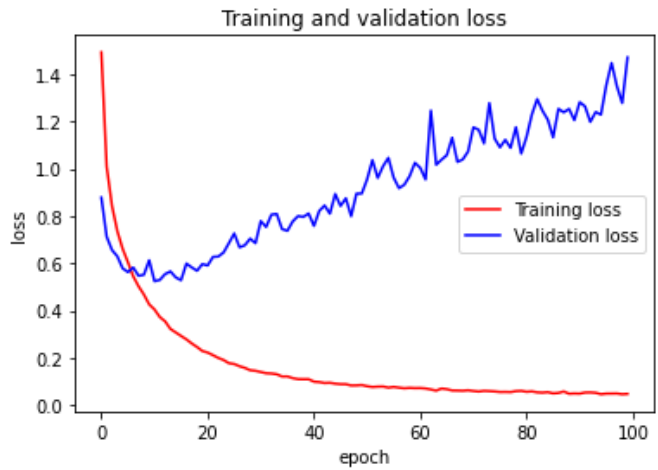

**Figure 5 :** *Model Loss*

**f(S+R)1 loss and accuracy**

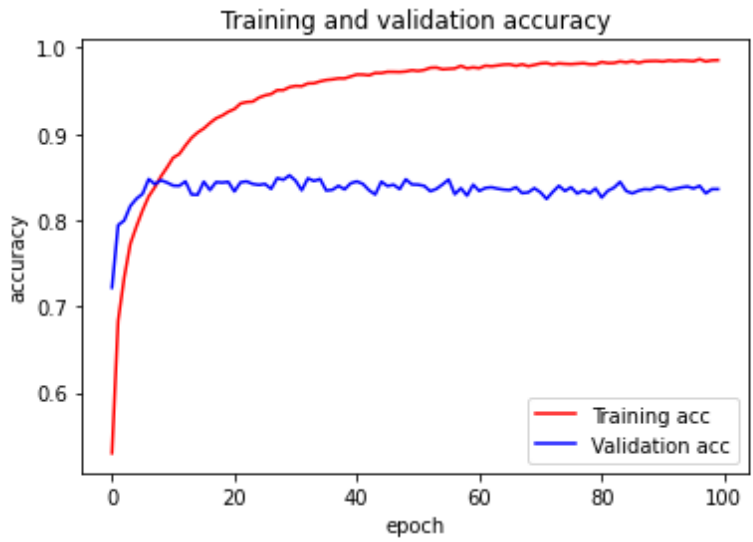

*Figure 6: Model Accuracy*

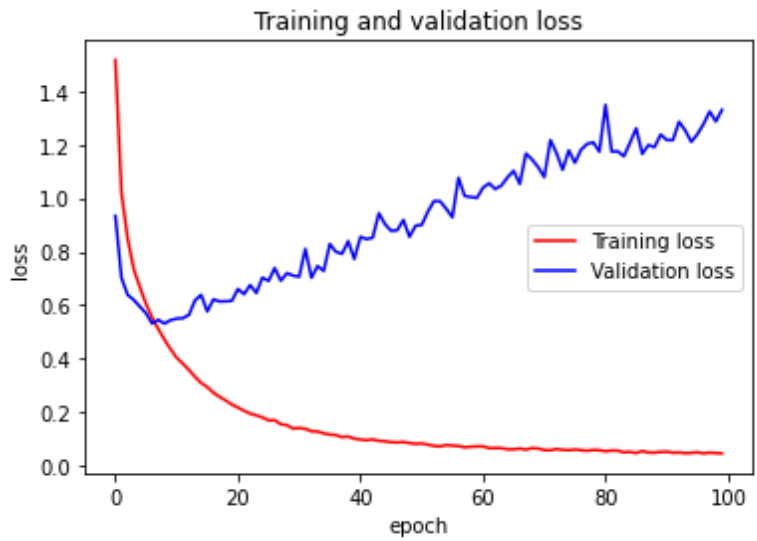

*Figure 7: Model Accuracy*

**f<sub>R</sub> loss and accuracy**

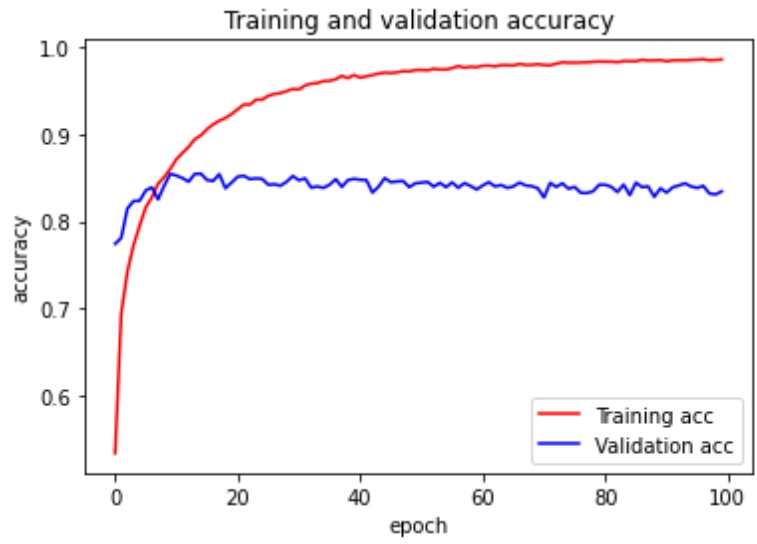

*Figure 8:* *Model Accuracy*

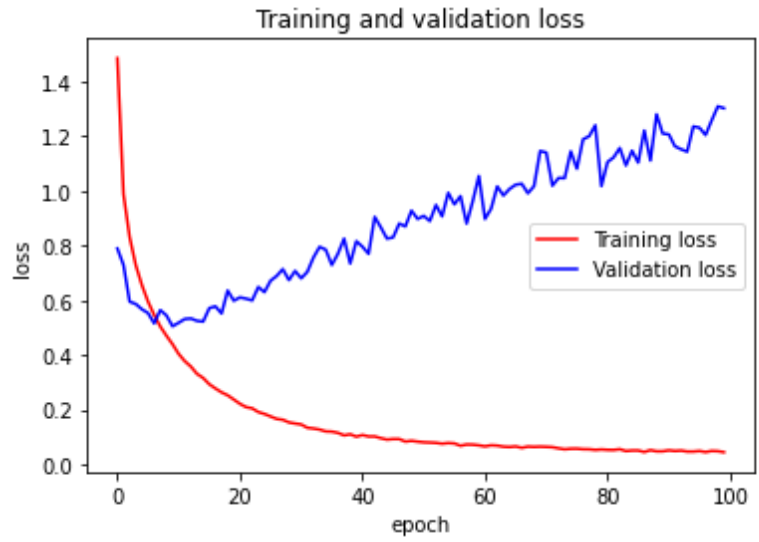

*Figure 9:* *Model Accuracy*